\documentclass{ieeeaccess}
\usepackage{mathptmx}
\usepackage[scaled]{helvet}

\usepackage{amsmath,amssymb,amsfonts}
\usepackage{algorithmic}
\usepackage{graphicx}
\usepackage{textcomp}

\usepackage{bm}
\usepackage{multirow}
\usepackage{siunitx}
\usepackage{hyperref}
\makeatletter
\AtBeginDocument{\DeclareMathVersion{bold}
\SetSymbolFont{operators}{bold}{T1}{times}{b}{n}
\SetSymbolFont{NewLetters}{bold}{T1}{times}{b}{it}
\SetMathAlphabet{\mathrm}{bold}{T1}{times}{b}{n}
\SetMathAlphabet{\mathit}{bold}{T1}{times}{b}{it}
\SetMathAlphabet{\mathbf}{bold}{T1}{times}{b}{n}
\SetMathAlphabet{\mathtt}{bold}{OT1}{pcr}{b}{n}
\SetSymbolFont{symbols}{bold}{OMS}{cmsy}{b}{n}
\renewcommand\boldmath{\@nomath\boldmath\mathversion{bold}}}
\makeatother

\def\BibTeX{{\rm B\kern-.05em{\sc i\kern-.025em b}\kern-.08em
    T\kern-.1667em\lower.7ex\hbox{E}\kern-.125emX}}
\begin{document}
\history{Date of publication xxxx 00, 0000, date of current version xxxx 00, 0000.}
\doi{10.1109/ACCESS.2017.DOI}

\title{AICRN: Attention-Integrated Convolutional Residual Network for Interpretable Electrocardiogram Analysis}
\author{\uppercase{
J. M. I. H. Jayakody}\authorrefmark{1},
A. M. H. H. Alahakoon\authorrefmark{1}, 
C. R. M. Perera\authorrefmark{1}, 
R. M. L. C. Srimal\authorrefmark{1}, 
Roshan Ragel\authorrefmark{1}, 
Vajira Thambawita\authorrefmark{2} and
Isuru Nawinne\authorrefmark{1}}

\address[1]{Department of Computer Engineering, University of Peradeniya, Peradeniya 20400, Sri Lanka}
\address[2] {Simula Research Laboratory, Kristian Augusts Gate 23, 0164 Oslo}

\markboth
{J. M. I. H. Jayakody \headeretal: AICRN: Attention-Integrated Convolutional Residual Network for Interpretable Electrocardiogram Analysis}
{J. M. I. H. Jayakody \headeretal: AICRN: Attention-Integrated Convolutional Residual Network for Interpretable Electrocardiogram Analysis}

\corresp{Corresponding author: Roshan Ragel (e-mail: roshanr@eng.pdn.ac.lk)}

\begin{abstract}
The paradigm of electrocardiogram (ECG) analysis has evolved into real-time digital analysis, facilitated by artificial intelligence (AI) and machine learning (ML), which has improved the diagnostic precision and predictive capacity of cardiac diseases. This work proposes a novel deep learning (DL) architecture called the attention-integrated convolutional residual network (AICRN) to regress key ECG parameters such as the PR interval, the QT interval, the QRS duration, the heart rate, the peak amplitude of the R wave, and the amplitude of the T wave for interpretable ECG analysis. Our architecture is specially designed with spatial and channel attention-related mechanisms to address the type and spatial location of the ECG features for regression. The models employ a convolutional residual network to address vanishing and exploding gradient problems. The designed system addresses traditional analysis challenges, such as loss of focus due to human errors, and facilitates the fast and easy detection of cardiac events, thereby reducing the manual efforts required to solve analysis tasks. AICRN models outperform existing models in parameter regression with higher precision. This work demonstrates that DL can play a crucial role in the interpretability and precision of ECG analysis, opening up new clinical applications for cardiac monitoring and management.
\end{abstract}


\titlepgskip=-15pt

\maketitle

\section{Advancements in ECG Diagnostic Methods}
\label{sec:Background}
\PARstart{E}{lectrocardiogram} (ECG) utility has been greatly enhanced over the years. From 1960 \cite{b1}, the utility of ECG improved with technological advancements. During the last 65 years, ECG analysis has evolved from interpretation on paper to a sophisticated digital analysis with the help of AI and ML to improve predictive accuracy for cardiac conditions. ECG analysis is an essential tool for medical professionals in cardiac diagnostic procedures.

\begin{figure}[ht]
    \centering
    \includegraphics[width=3in,height=3in,clip,keepaspectratio]{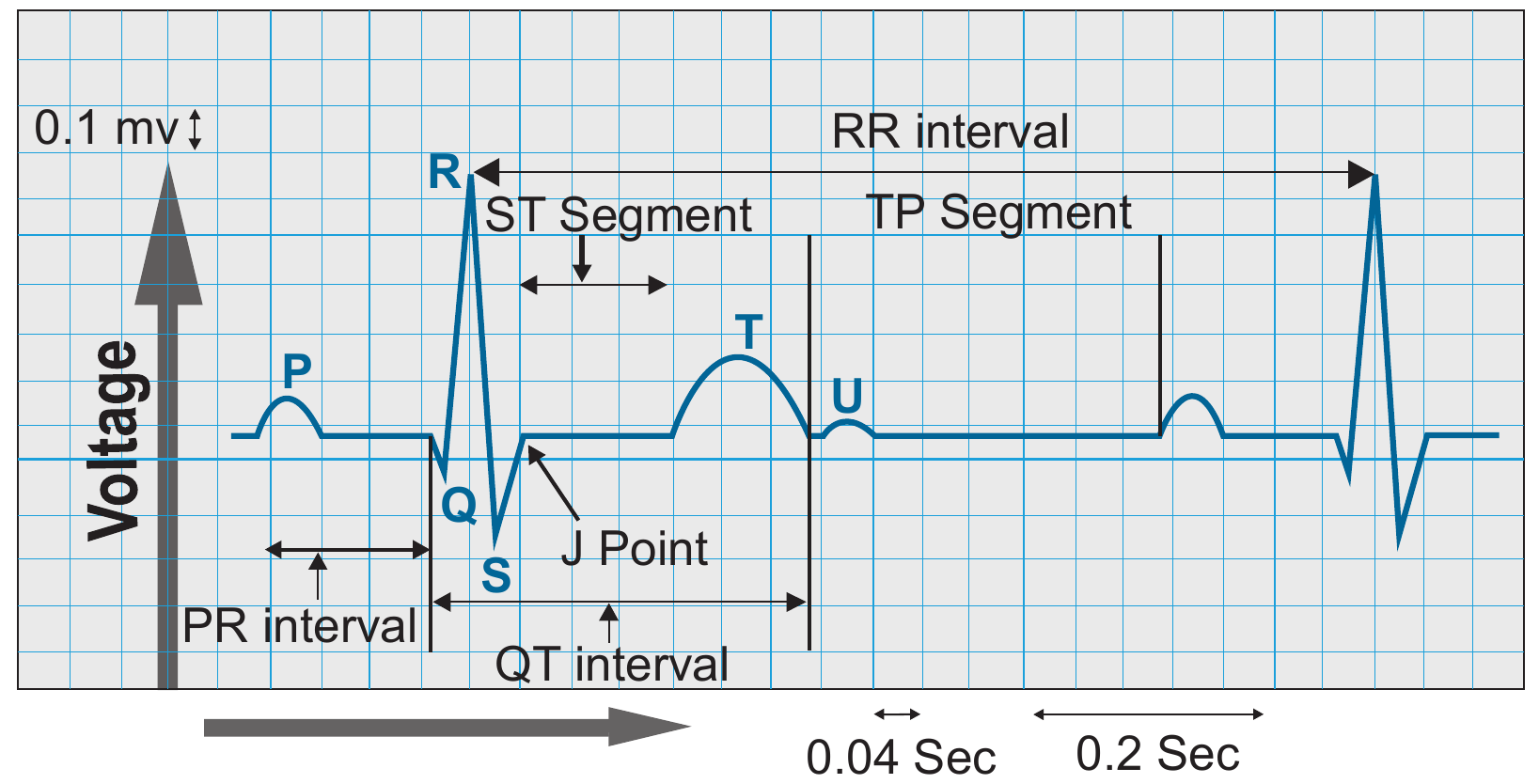}
    \caption{ The ECG grid is used to evaluate heart rhythms by measuring electrical activity over time. This structured form of an ECG is critical to assess heart rate, rhythm, conduction delays, and associated abnormalities of the heart and can be used to diagnose arrhythmias, ischemia, and conduction disorders \cite{b2}.\label{fig:ecgsignal}}

\end{figure}

\begin{figure}[ht]
    \centering
    \includegraphics[width=3in,height=3in,clip,keepaspectratio]{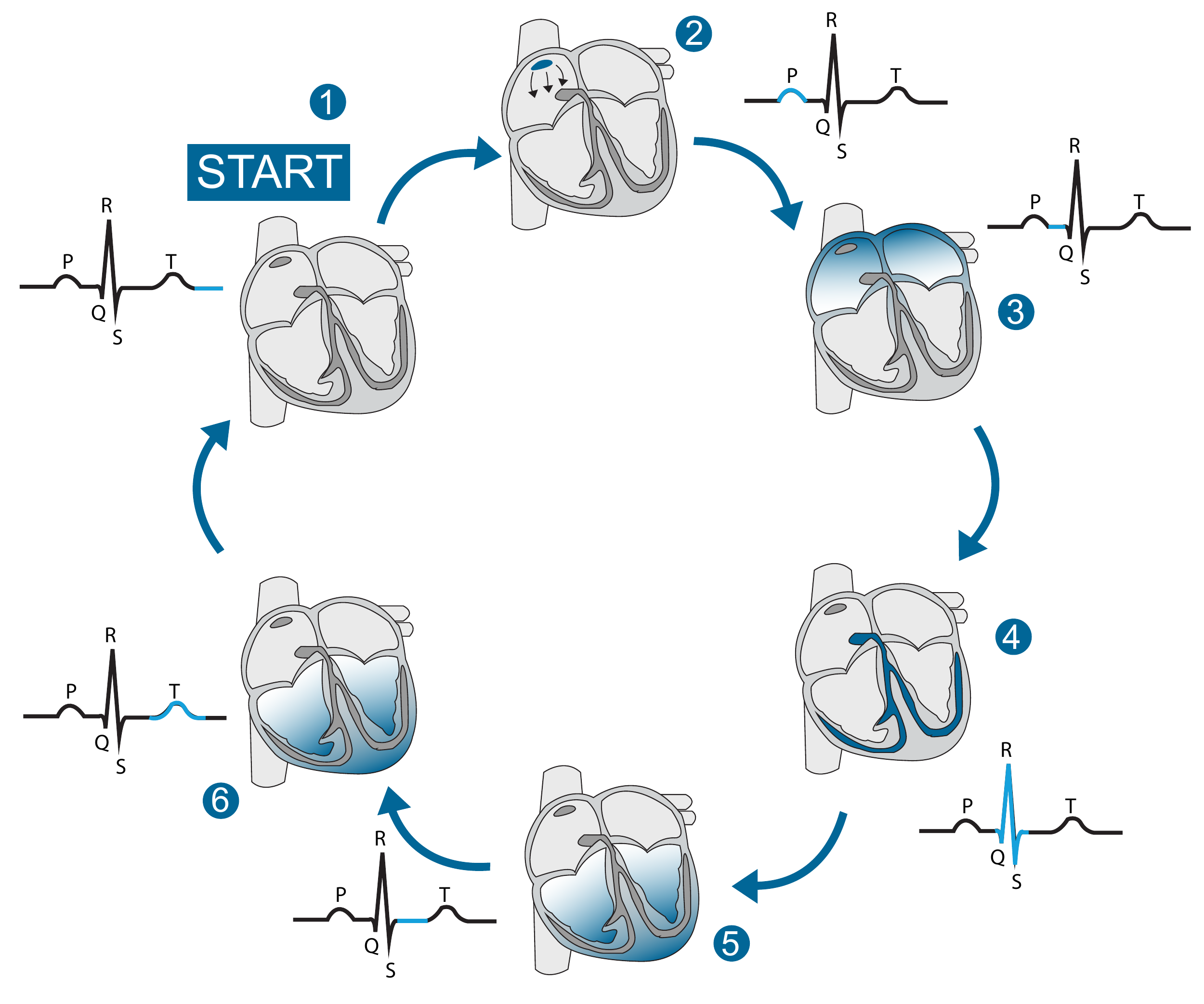}
    \caption{The cardiovascular cycle begins with atrial depolarization (P wave) when the sinoatrial node fires. This causes contraction in the atrium, and blood enters the ventricles. A slight delay at the atrioventricular node (PR segment) prevents some filling of the ventricles before depolarization (Q wave). The electrical impulse transmigrates through the bundle of His and Purkinje fibers before it reaches the ventricles and causes them to contract (QRS complex), increasing blood flow to the aorta and pulmonary artery. This increase in blood flow causes the ventricles to relax at the same time as the ST segment, a period preceding full repolarization (the T wave), followed by full recovery of cardiac activity \cite{b2}.
    }
    \label{fig:heartcycle}
\end{figure}

Twelve leads are placed on the thorax and limbs, and 10 electrodes are used to measure the electrical voltage activity of the heart. Electrodes attached to the body via leads detect electrical signals from the heart during depolarization and repolarization, transmitting them to a recording device. The voltage-time graph (Figure~\ref{fig:ecgsignal}) shows specific waves and intervals of cardiac conduction during different stages of the cardiac cycle (Figure~\ref{fig:heartcycle}). The ECG shows major components such as the P wave, QRS complex, and T wave, which indicate different electrical activities in the heart. Since the heart has its own conduction system (originating from the sinoatrial node), the ECG has become an essential tool for diagnosing cardiovascular diseases. Interpreting an ECG requires understanding both normal and abnormal waveforms and their correlation with the patient’s clinical symptoms and medical history.

Intervals, amplitudes, and heart rate are vital ECG parameters to evaluate heart functions such as conduction, repolarization, and overall performance. The QT interval gauges the recovery of the ventricles from depolarization, which is crucial for muscle contraction. Prolonged QT intervals indicate the risk of arrhythmias, especially those altering ECG waves. The PR interval measures the time for electrical signals from the atria to the ventricles. Extended PR intervals can signal a heart block of the first degree, while progressive lengthening with dropped beats suggests a second-degree or complete heart block \cite{b3}. The duration of the QRS complex assesses the propagation of the electrical signal through the lower chambers of the heart. Widened QRS complexes can indicate pathway delays or obstructions, with certain arrhythmias, such as rapid lower chamber heart rates, having distinct QRS patterns. Taking a look at the shape and structure of the QRS complex in an ECG will help pinpoint an arrhythmia with its severity.

Heart rate represents the cardiac function of how frequently the heart beats, reflecting the heart's ability to pump blood throughout the body. Heart rate monitoring offers excellent information on various functions of the heart. An abnormally slow or fast heart rate could lead to different cardiac conditions. The peak amplitude of the R wave shows the electrical voltage of the ventricular depolarization. The strength of the signal is proportional to the strength of the ventricles that contract to pump blood. It also tells about the mass and condition of the ventricular muscle. The RR interval tells the time between consecutive peaks of the R wave, which implies electrical stability and intrinsic rhythm. The amplitude of the T wave tells us about ventricular repolarization, which is the process of restarting cardiac muscle cells as needed to start the next heartbeat. These parameters are valuable in diagnosing various cardiac conditions, developing effective treatment plans tailored to individual patients, and accurately measuring the therapeutic effects of medications over time.

Computerized electrocardiography systems automatically perform ECG analysis and interpretation to enable a precise diagnosis with timely intervention using continuous ECG monitoring, detecting small irregularities for further evaluation by healthcare professionals. Deep learning (DL) for ECG analysis extends traditional cardiac diagnosis to those in which systems recognize emotional states or biometric identification using ECG for novel applications. They also enable the storage, retrieval, and sharing of ECG data, facilitating remote consultations and long-term monitoring. These reasons support the development of continuous monitoring using DL models optimized for low power consumption and high efficiency that promise to improve patient care outside traditional clinical settings \cite{b4}.

\section{Deep Learning Approaches for ECG Analysis}
\label{sec:introduction}
ECG is widely used in medical procedures due to its insight and affordability. Experienced healthcare professionals analyze ECG results to diagnose and monitor cardiac conditions, preparing treatment plans. Integrating technology into ECG analysis increases diagnostic and monitoring accuracy, making faster and informed decisions that improve patient outcomes. AI aids ECG analysis with DL algorithms that learn from large datasets, identifying cardiac functions precisely and rapidly to improve diagnostic and monitoring accuracy in patient care. ECG analysis data sets include raw signals, cardiologist annotations, patient metadata, longitudinal data that track cardiac health over time, and multimodal data that combine ECG with other health indicators, which are crucial to train accurate AI models. Deep ECG analysis is vital for developing interpretable models that expand knowledge and apply beyond traditional clinical settings.

Traditional ECG analysis faces a wide range of challenges. It relies heavily on the experience and expertise of the clinician, which can lead to variability in diagnosis. Traditional ECG analysis is based on experience and knowledge. Experience and knowledge can lead to significant differences in diagnosis as a result of human judgment or personal bias. Additionally, ECG analysis requires manual review and interpretation of the tracings. The time required to review and interpret ECGs affects cautious decision making within an urgent care setting. Standardization of ECG diagnosis is time-consuming and costly without computerized tools. Even with advanced tools, subtle ECG issues can be missed due to noise, overlapping conditions, or interpretation challenges.

Recent discoveries highlight DL's superiority over traditional methods in medical information analysis, especially in the evolving subdomain of ECG analysis. The main reason is the intriguing and diverse analytical potential of DL for ECG analysis. DL models identify complex patterns in the data, detecting subtle, non-obvious cardiac abnormalities in ECG signals with high precision, reducing misdiagnosis rates. DL processes ECG data in real time, offering immediate feedback and diagnoses, crucial in emergency settings for quick life-saving decisions. Unlike traditional methods that require manual feature identification and annotation, DL algorithms learn important features from raw data, boosting diagnostic efficiency. DL handles vast data, which is vital for healthcare's growing datasets, enabling robust statistical analysis and model training. DL serves as a diagnosis tool for future cardiac events by analyzing trends and patterns in ECG data over time, helping with preventive care and long-term health monitoring. Learning from diverse datasets, DL models recognize patient-specific patterns, leading to personalized diagnosis and treatment plans. Integrating DL in ECG analysis addresses traditional challenges \cite{b5}, signaling a shift to autonomous real-time cardiac care monitoring systems.

Various DL algorithms, such as transformer models, have been studied for ECG diagnosis. Successful studies have shown that they capture complex short-term ECG patterns \cite{b6}, \cite{b7}, \cite{b8}, \cite{b9}. Transformer models are computationally efficient and scalable. The adoption of CNNs makes it feasible to detect abnormalities/arrhythmias by mixing spatial and temporal analysis \cite{b10}, \cite{b7}. Research has also been done on DL algorithms for automatically finding and labeling parts of an ECG signal, such as peaks, and detecting arrhythmias \cite{b11}, \cite{b12}, \cite{b13}, \cite{b14}, \cite{b15}. Studies such as \cite{b16} proposed a DL approach for the identification of the QRS complex using multiple leads. Their performance matches that of human expert systems. Models such as \cite{b6} predict all heartbeats within a single segment. They are compact, end-to-end solutions for the detection of arrhythmias. Architectures like \cite{b17} and \cite{b18} have been developed to predict risk, age, and sex based on standard ECG parameters. These architectures are low-cost and non-invasive methods for the risk assessment of non-critical ECG findings.

In addition, the clinical applications of DL models in ECG analysis have several challenges, including heterogeneity and interpretability of the data. The generalizability of the models should be determined through rigorous testing in a wide range of patients \cite{b19}. Explainability is also essential to gain the trust of clinicians \cite{b20}, \cite{b21}, \cite{b22}, \cite{b23}. Techniques such as ECGradCAM (which uses attention maps to provide further clarification and detail into decision-making processes in deep neural networks) are important to improve model transparency \cite{b23}, \cite{b24}, \cite{b25}. Integrating DL models into ECG analysis is a major trend that has reshaped cardiac care. It will improve diagnostic accuracy, efficiency, and implement real-time autonomous monitoring. As technology develops, it can be integrated into clinical workflows and potentially improve cardiac care outcomes in various settings. The continued evolution of DL models and their ECG applications shows its dynamic nature, from the diagnosis and management of cardiovascular diseases \cite{b26}, \cite{b27}, \cite{b28}, \cite{b29} to wearable devices and emotion identification systems, which have broader applications beyond the traditional clinical settings \cite{b30}, \cite{b31}, \cite{b32}.

The paper employs advanced deep learning techniques to automate ECG analysis, significantly enhancing precision in interpretable cardiac diagnostics. There are many studies on identifying and categorizing ECG characteristics. However, they lack interpretability. Direct prediction of specific ECG measurements of major intervals and amplitudes allows for a detailed and advanced interpretation of cardiac health. This provides a deeper and individualized understanding of cardiac features and conditions \cite{b23} instead of simplifying or generalizing the analysis by grouping the data into smaller, less specific categories.

This study is based on several core research questions. First, can deep learning models be utilized to accurately regress critical electrocardiogram (ECG) parameters? Second, how much can the integration of attention mechanisms within a convolutional residual network architecture (CRN) enhance the precision and interpretability of ECG analysis compared to conventional methods? Third, can an automated deep learning-based system effectively support real-time cardiac monitoring, reduce human diagnostic errors, and consistently outperform existing state-of-the-art models across diverse clinical settings?

The main purpose of this work is to develop a novel deep learning architecture called the attention-integrated convolutional residual network (AICRN), which is applied to the regression of the ECG parameters. By integrating attention mechanisms into a residual network, the purpose of this research is to significantly improve the precision, efficiency, and clinical interpretability of ECG diagnostics through the designed architecture. In addition, integrating this architecture into cardiac analysis automation can improve the precision of cardiac diagnosis, reduce manual intervention needs, and facilitate continuous real-time monitoring of cardiac activity to improve patient outcomes. As an example, the prediction of patient responses a few minutes earlier in an ongoing cardiac surgery and the monitoring of the cardiac effects of the cardiovascular system that affect drugs. Furthermore, this work can be improved or integrated into explainable systems that provide greater transparency and interpretability, ultimately making them more understandable, reliable, and acceptable to medical professionals to accurately analyze and interpret cardiac activities.


The main contributions of this study are as follows.

\begin{itemize}
    \item \textbf{Novel ECG Regression Architecture:} We developed a novel attention-integrated ResNet-based convolution neural network architecture optimized for regressing key ECG parameters, which is valuable in interpretable ECG analysis.

    \item \textbf{Open-Source Implementation:} We introduce an ECG monitoring software capable of continuously tracking ECG parameters over time, reducing the time required for manual ECG analysis and improving clinical decision-making and medical outcomes. We release our trained models, the complete source code of the application, and all data processing scripts as open source resources. This facilitates reproducibility and encourages future research and practical applications within the community. The resources are available at \href{https://github.com/cepdnaclk/e17-4yp-Comprehensive-ECG-analysis-with-Deep-Learning-on-GPU-accelerators}{our GitHub repository}.
\end{itemize}

\section{Methods}
\label{sec:Methods}
In this study, the main objectives include i. Design and implementation of the DL architecture for the regression of critical ECG parameters; ii. Training and validation of the models using the PTB-XL dataset for the study of robustness and clinical relevance; iii. Performing rigorous evaluations on the performance of these DL models, insofar as they compare with previous well-established deep learning model architectures such as ResNet, QTNet and QTGAN by several metrics such as the mean absolute error (MAE), the root mean square error (RMSE) and the coefficient of determination (R2 score). iv. Practical implementation of ECG monitoring for real-time diagnosis and management of cardiac conditions in clinical and remote health care systems.

\subsection{Data Preparation}

The PTB-XL dataset is a large publicly available ECG dataset with an extensive collection of individual annotated ECG recordings. It is widely used in the research and development of specific ECG analysis techniques such as ML and DL models. Additionally, it offers comprehensive data with specific characteristics of the population, the specific characteristics of tissue death due to the absence of blood supply, and the likelihoods of diagnoses, which are important resources for research in this area \cite{b33}.

These characteristics are crucial in ensuring that research is based on realistic and clinically relevant scenarios. \cite{b34} offers a comprehensive evaluation of DL models in ECG analysis using the PTB-XL data set. These insights from the literature provide a solid foundation for choosing model architectures and evaluation metrics, ensuring that methods are innovative and aligned with current best practices. The study benchmarks architectures such as ResNets and inception-based models in tasks that resemble the prediction of the ECG statement, the estimation of age and sex, and the evaluation of signal quality. The findings show that DL models, including ResNets, use complex convolutional layers for efficient information processing, outperforming feature-based algorithms in ECG analysis. It also explores model interpretability, uncertainty estimation, and transfer learning, offering valuable insights for robust and accurate ECG analysis systems. The wide range of uses of this data set in cardiac research suggests its importance as a benchmark for the evaluation and improvement of current ECG analysis methods.

The PTB-XL ECG dataset is a large-scale source of clinical 12-lead ECG recordings from 18,869 patients. Cardiologists annotate these ECGs according to the SCP-ECG standard, with 71 evaluation criteria. The ethical implications arise because this dataset contains human medical information. The ethical guidelines for the use of this material should be carefully examined before ethically adopting them and concerning confidentiality. The data set contains default training and test splits, which will allow the development of DL algorithms to automatically interpret a series of ECGs in a single training set. Therefore, a systematic approach to ECG data management for DL models (particularly in flexible, efficient frameworks) is essential to train robust models on medical datasets such as PTB-XL. Predefined training, validation, and testing splitting allows systematic algorithm evaluation and can allow robust comparisons to the optimal model and reproducibility. This study stresses the experimental setting in the selection of the PTB-XL dataset due to its diversity and the large annotation corpus, which allows detailed analysis of cardiac conditions. Structured data enables exhaustive statistical and computational tasks to ensure high levels of significance and usefulness in clinical settings such as emergencies and surgeries.

\subsubsection{Data Preprocessing}
In this study of ECG analysis, the PTB-XL data set was meticulously prepared to train DL models. Important settings were set for the ECG data and organized in a logical way. The evaluations were performed according to the best training performance of the models. Incomplete records were removed. Data normalization was performed. Unnecessary data parts were eliminated. The ECG data preparation step is the fundamental aspect that determines the reliability of regression modeling. This step is required because only a perfect configuration of the ECG data influences the regression modeling ability.

At the start of data preprocessing, parameters such as ECG data types (PR, QT, QRS intervals, calculated heart rate using the RR interval, R wave peak amplitude, and T wave amplitude) are determined. The data split ratios and the size of the data matrix are defined for each subset of datasets. This list of data vectors is organized to allow data manipulation and analysis. The data is divided into training, validation, and testing subsets, which are randomized to maximize the number of replicates for each dataset. The data is then cleaned, which is based on missing values in the major columns, to ensure the completeness and accuracy of the data. As there is scaling support for the normalization of ECG data, this process is done to efficiently train neural networks. These steps are based on the best neural network training data handling strategy according to the neural network architecture principles, observed in \cite{b35}. The data of the ECG signal itself is processed by removing unnecessary channels, such as III, AVR, AVL, and AVF, leaving only key channels such as I, II, and V1 to V6. These ECG leads are strategically selected as input features for parameter regression because they provide comprehensive views of the heart's major regions, which are important for the selected parameter regression. In this process, a significant amount of computational load is avoided, since only information relevant for the selected parameters' regression is included in the models' training. Lastly, processed ECG data undergo a renormalization step, during which time they are systematically transformed into tensor structures that align with the required input format. This transposition ensures that the data is accurately reshaped and presented in the correct dimensions, aligning precisely with the structural requirements and expectations of the subsequent stages of the analysis, thereby facilitating a smoother and more reliable flow of information throughout the processing pipeline.

\begin{figure}[ht]
    \centering
    \includegraphics[width=2.2in,height=2.2in,clip,keepaspectratio]{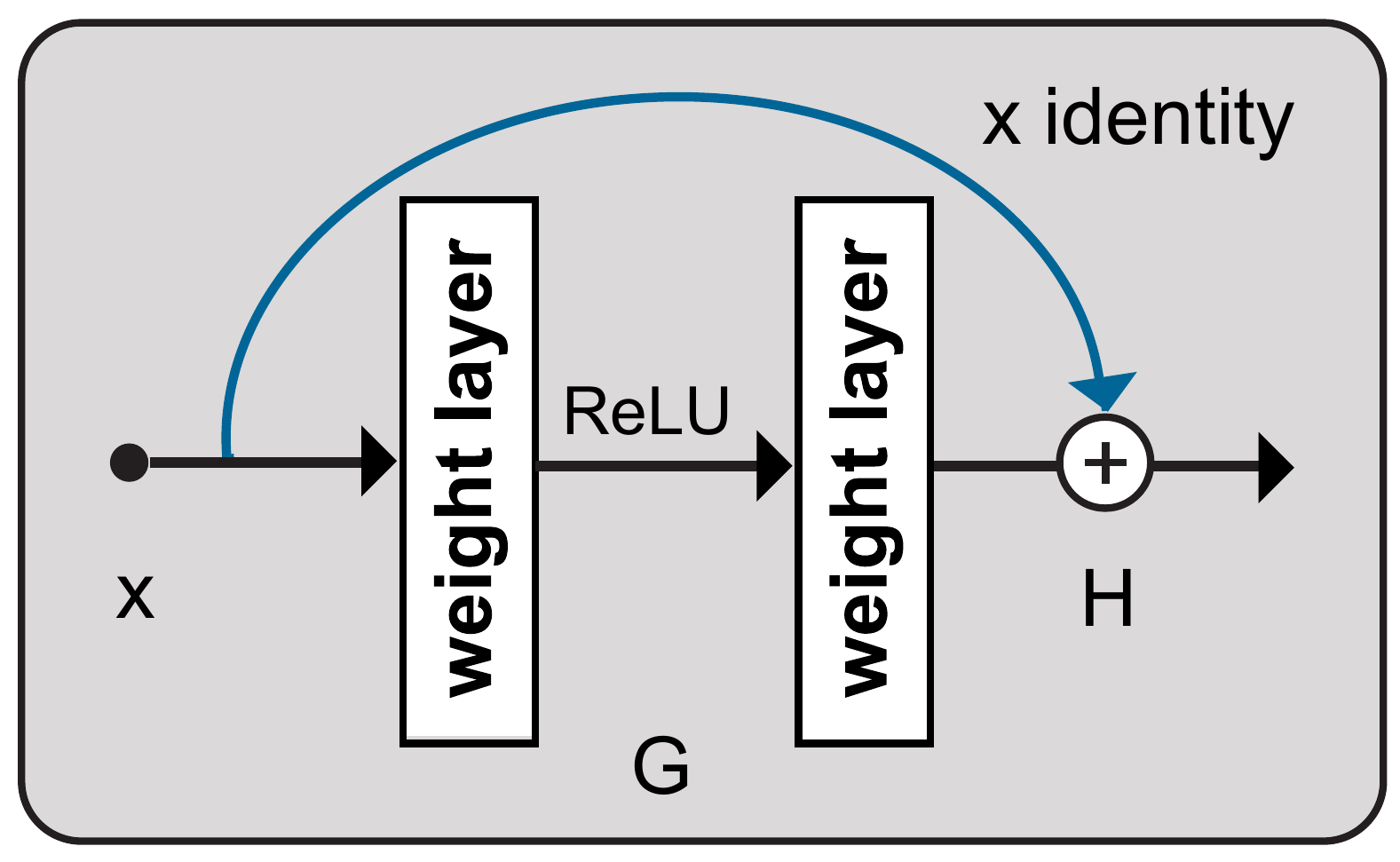}
    \caption{ Residual Network (ResNet) block, specifically showcasing an identity shortcut connection. The diagram depicts input x passing through two sequential operations: a weighted layer followed by a ReLU activation, collectively referred to as G. Simultaneously, x bypasses these operations via an identity shortcut, allowing the original input to be added directly to the output of the weighted layer and ReLU activation. The outputs from both pathways merge at a summation point, resulting in the final output G(x) + x, which is then forwarded \cite{b39}.
}
    \label{fig:residualblock}
\end{figure}

\subsection{Deep Learning Architecture}
The regression study uses clinically relevant AICRN measures, such as the intervals, amplitude, and heart rate of the eight selected leads, leveraging the PTB-XL data set with extensive testing and optimizations. This configuration is designed based on its proven efficacy in handling similar data-rich environments, a choice supported by recent advances in DL \cite{b23}, \cite{b36}, \cite{b37}, \cite{b38}. Target labels such as PR interval, QRS duration, heart rate, QT interval, R wave peak amplitude, and T wave amplitude are iterated over in a loop to train the models for each parameter. In transitions to the model's operational dynamics, these models employ attention mechanisms along with convolutional residual modules, ideal for time-series data, dynamically focusing on important and subtle features of the input data.

\begin{figure}[ht]
    \centering
    \includegraphics[width=3.4in,height=3.4in,clip,keepaspectratio]{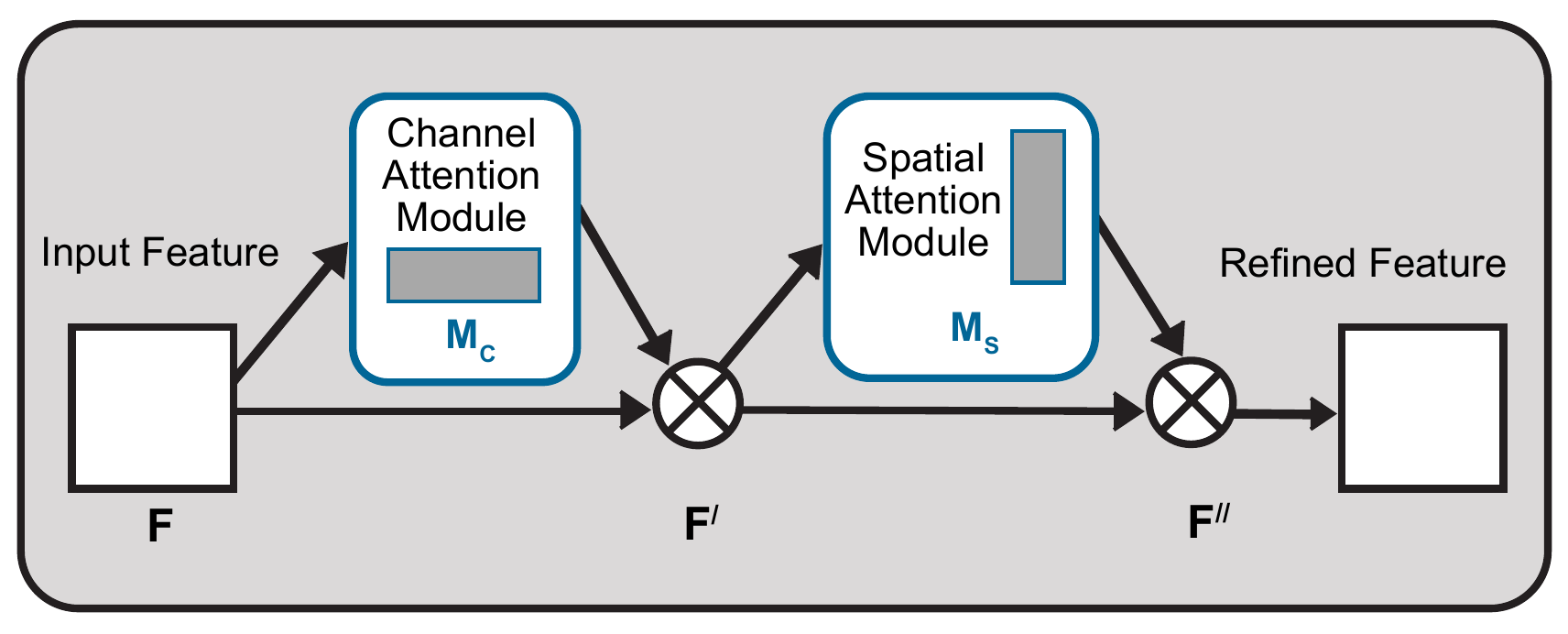}
    \caption{ Schematic diagram of a CBAM. It shows two primary sections: one for channel attention and one for spatial attention. Each section depicts inputs and outputs with arrows indicating the flow of data. The channel attention part processes a feature map through mechanisms like average pooling and max pooling to prioritize certain channels. Subsequently, the spatial attention section further refines the attention by focusing on specific spatial features within the resulting channels, using a similar pooling strategy \cite{b40}.
}
    \label{fig:cbam}
\end{figure}

\subsubsection{Residual networks in processing ECG data}

Residual Network architectures, commonly known as ResNets, fundamentally improve the training of deep neural networks by addressing common challenges like vanishing and exploding gradients. These networks incorporate "shortcut connections" that allow some layers to be skipped, facilitating a smoother flow of gradients during training. In traditional networks, each layer aims to learn a direct mapping from input X to output H(X). ResNets, however, modify this approach by learning the difference, or residual, as shown in Equation~\ref{eq1}. This residual function is simpler to optimize because the network can set G(X) to zero if the identity mapping is optimal, which means Equation~\ref{eq2}.

\begin{equation}
G(X) = H(X) - X
\label{eq1}
\end{equation}

\begin{equation}
H(X) = X
\label{eq2}
\end{equation}

The structure shown in Figure~\ref{fig:residualblock} illustrates a basic building block of a ResNet. Here $\oplus$ denotes element-wise addition. The input X is processed through two weight layers separated by a ReLU (Rectified Linear Unit) activation function. The output of this sequence, G(X), is then added back to the original input X, before being passed through another ReLU activation. This step allows the network to either learn a useful new transformation or effectively pass the input directly through if that provides a better gradient path. These identity shortcuts ensure that the depth of the network increases without significant increases in computational complexity. This capability allows ResNets to learn increasingly complex features and achieve performance in ECG analysis.

\begin{figure}[ht]
    \centering
    \includegraphics[width=3.3in,height=3.3in,clip,keepaspectratio]{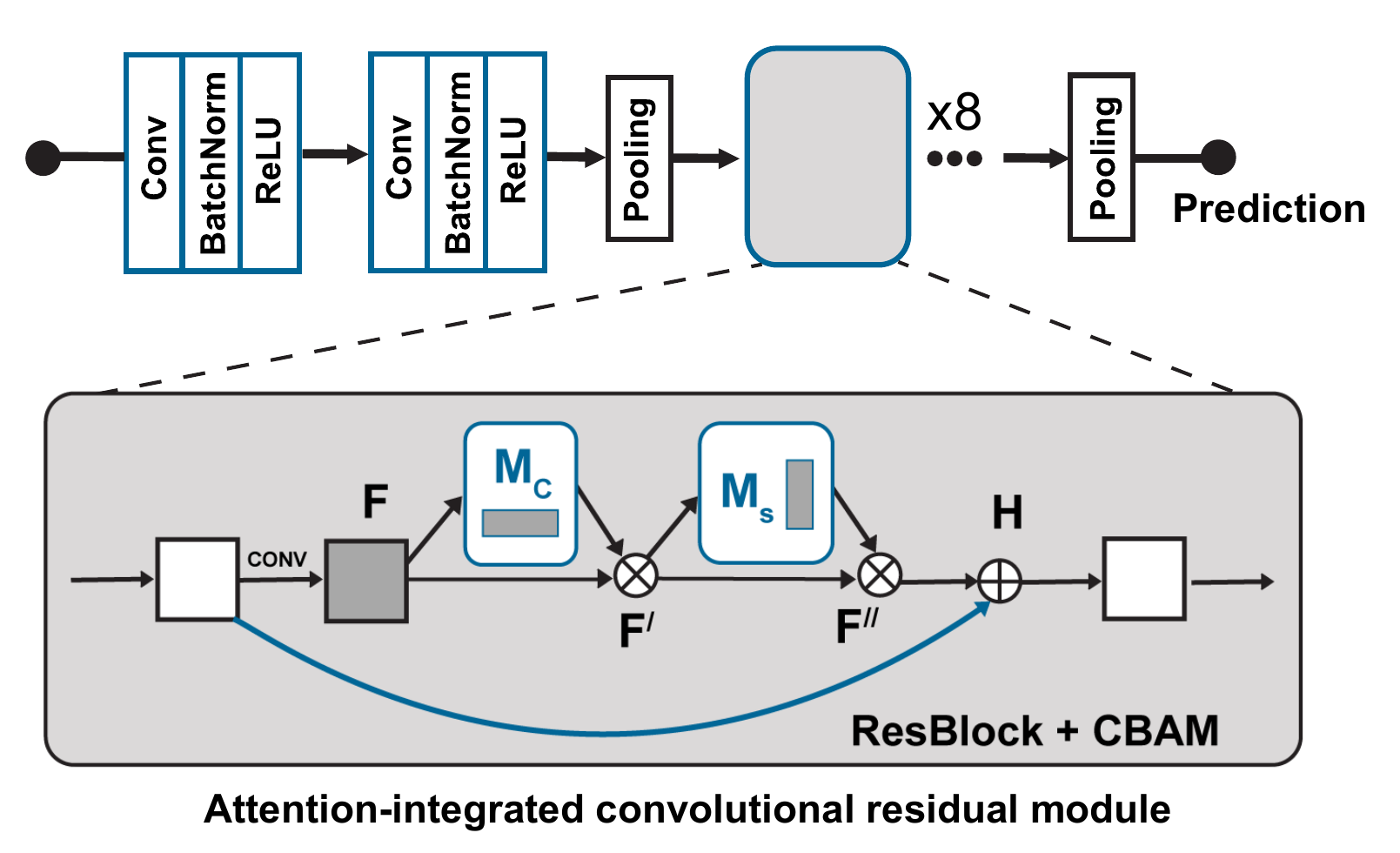}
    \caption{ The method uses attention-integrated convolutional residual network-based architecture for regressing every ECG parameter. This architecture uses convolutional, batch normalization, ReLU activation, and average pooling stages. Besides these base building blocks, the DL architecture consists of eight attention-integrated convolutional residual modules (grey blocks) composed of two sequential convolution blocks, batch normalization, and ReLU activation.
    }
    \label{fig:model}
\end{figure}

\subsubsection{Attention mechanisms in processing ECG data}
The designed neural network architecture has residual CNN blocks with large kernels designed to capture ECG characteristics, global clustering for predictions, and convolutional block attention modules (CBAM) \cite{b40} shown in Figure~\ref{fig:cbam}. These components provide simple yet effective attention mechanisms appropriate for regressing intervals, amplitudes, and heart rates, highlighting the role of each element in enhancing models' performance. It enhances CNN capabilities by focusing on important features within an ECG dataset through two processes. They are determining 'what' features to emphasize using the channel attention module (CAM) and identifying 'where' these important features are located using the spatial attention module (SAM). CAM allows the network to decide which features in the data set are significant. This is done through two main steps. The spatial squeeze step reduces the spatial dimensions of the feature map to highlight important features across different channels. It uses two pooling operations. The average pooling (AvgPool) operation in Equation~\ref{eq3} calculates the average value of elements in a specific region of the feature map, helping to understand the general presence of features. The max pooling (MaxPool) operation in Equation~\ref{eq4} picks the maximum value within a region of the feature map, highlighting the most prominent feature. They transform the original feature map F into two new feature descriptors.

\begin{equation}
F_{c_{\text{avg}}} = \text{AvgPool}(F)
\label{eq3}
\end{equation}

\begin{equation}
F_{c_{\text{max}}} = \text{MaxPool}(F)
\label{eq4}
\end{equation}

Both descriptors from the spatial squeeze step are processed using a shared Multi-Layer Perceptron (MLP), a type of small neural network. The MLP outputs are combined and passed through a sigmoid function shown in Equation~\ref{eq5} to scale the results between 0 (least important) and 1 (most important). Here, \( \sigma \). denotes the sigmoid function that helps normalize the outputs, ensuring that they are between 0 and 1.

\begin{equation}
M_c(F) = \sigma(\text{MLP}(F_{c_{\text{avg}}}) + \text{MLP}(F_{c_{\text{max}}}))
\label{eq5}
\end{equation}

SAM complements the CAM by focusing on 'where' the significant features are spatially located within the ECG signals. It also consists of two steps. The channel squeeze is similar to the spatial squeeze but applies across the channel dimension. This step highlights which parts of the ECG signal are most informative by applying the same two pooling operations. The spatial feature descriptor outputs of the channel squeeze are concatenated and then processed by a convolution layer with a 7x7 filter size, followed by a sigmoid function to generate the spatial attention map in Equation~\ref{eq6}.

\begin{equation}
M_s(F) = \sigma(f_{7 \times 7}([F_{s_{\text{avg}}}; F_{s_{\text{max}}}]))
\label{eq6}
\end{equation}

Here, \( f_{7 \times 7} \) represents the convolution operation, and \( \sigma \) is the sigmoid function normalizing the outputs. The output attention maps of CAM and SAM are applied to the input feature map in sequence. First, the feature map \( \mathbf{F} \) is refined by the channel attention map Equation~\ref{eq7}. Then, this channel-refined feature map \( \mathbf{F'} \) is further refined by the spatial attention map in Equation~\ref{eq8}. Here $\otimes$ denotes element-wise multiplication.

\begin{equation}
F' = M_c(F) \otimes F
\label{eq7}
\end{equation}

\begin{equation}
F'' = M_s(F') \otimes F'
\label{eq8}
\end{equation}

By dynamically adjusting the focus on the most informative features, CBAM enhances the representation power of CNNs, as shown in Figure~\ref{fig:cbam}. This module is versatile and efficient. CBAM is designed to be lightweight and can be easily integrated into any residual CNN architecture without significant computational overhead. CBAM refines feature attention based on both 'what' and 'where', while residual modules ensure smoother gradient flow, together enhancing performance in complex ECG analysis.

\subsubsection{Detailed Configuration of NN architecture}

The first convolutional layer transforms the input feature maps from the number of input channels to the number of filters, applying batch normalization to stabilize and accelerate training. The second convolutional layer then processes the feature maps from the first layer's number of filters to those of the second layer. After each batch normalization, it applies a Leaky ReLU with a negative slope coefficient of 0.1 to introduce non-linearity and effectively handle negative signals. Instead of the standard ReLU, it uses the Leaky ReLU activation function to prevent the issue of dying neurons, which permanently output zeros due to negative inputs. It uses two sequential convolutional layers to deepen the network, enabling sophisticated feature extraction.

Batch normalization reduces internal covariate shifts, enhancing the training dynamics. Following two sequential convolutional layers, the data undergoes an initial transformation using average pooling with a kernel size of 2, effectively reducing the data dimensionality. A sequence of eight residual modules equipped with attention mechanisms processes the data sequentially (Figure~\ref{fig:model}). The attention-integrated convolutional residual module incorporates a residual structure with attention mechanisms (Figure~\ref{fig:model}), enhancing feature processing within deep neural networks. This module utilizes two convolutional layers with 'same' padding to maintain input and output dimensionality, ensuring stability throughout the learning process. The output of these layers is normalized to stabilize further learning. It uses dropout techniques in designing, setting a zero probability of 0.5 for input tensor elements to prevent overfitting and act as regularization. Global average pooling is applied to reduce feature dimensions to a single scalar per feature map. Finally, a linear layer maps the pooled features to the desired output size. This architecture optimizes performance by addressing overfitting and unstable training, enhancing accuracy and efficiency for complex tasks.

\subsection{Training and Evaluation}
\subsubsection{Training Setup}
Key training settings defined include an input shape of 8 for 8 LEADS and an output size of 1. Nadam optimizer learning rate was set to 0.0005 with 1000 epochs. These values were optimized from experiments, suggesting that they are the most appropriate ratio between training speed and model accuracy. This shows the fine-tuning of the models with the ECG data. The input shape is defined according to the ECG channels and time steps to capture the entire temporal dynamics of the electrical activity of the heart. Such a shape enhances the models' ability to learn from complex signal patterns. The training algorithm set target labels such as PR interval, QRS duration, heart rate, QT interval, peak amplitude of the R wave, and amplitude of the T wave, which are then iterated in a loop to train models for each parameter while ensuring focused and granular learning of parameters necessary for precise parameter regression. The batch size is set to 300 for both the training and validation data. This batch size is ideal for GPU memory cost savings and sufficient data throughput for stable gradient updates. Upon initialization with the designed attention-integrated residual architecture, the training process encompasses the entire training loop. This includes forward passes, backward passes for gradient calculation, weight updates, and validation runs to track performance across epochs. This comprehensive approach is critical for iterative refinement of the models, ensuring that each component contributes optimally to the final performance. During this loop, training progress is displayed so that debugging and monitoring of the training process can be performed. It also allows for feedback on how well your training is doing and identifies problems early.

\subsubsection{Validation Approach}
During training, each time the training epoch is closed, the system stops the training. This is important because it prevents overtraining and supports generalizability to new, unobserved data. Conversion of a validation loss to a score improves the performance. Improvement checks allow for a clear understanding of the progress made. An improvement with higher scores indicates greater model precision. The system checks to see if the current score is much better than the best score the system has seen, and if not, a counter increases. Training is stopped if the counter exceeds the patience limit. If the improved model accuracy is equal to or better than the best score, the top score is updated, and the counter is reset. This provides a dynamic feedback loop for constant monitoring and optimization of the models as they are trained, making training adjustments at the right time possible. Once a minimum level of validation loss is reached, the model state is stored on disk so that there is a possibility of recovering to the best state. Later in the training process, it is necessary. This guarantees the best-performing model during training and support for reproducibility since researchers have access to the best-performing configuration of the model during training. A saving condition is triggered when a new minimum is reached in validation loss, and the model weights are saved to a path marked as the best model so far. This will prevent overfitting and stop the training when further epochs do not improve the validation loss. This protects against declining returns on refined models.

\subsubsection{Performance Metrics}
In evaluating the predictive models' performance, this study chooses three main metrics. They are well-accepted in statistical analysis and provide a solid framework for evaluating the regression models' performance in DL research. These metrics are the Root Mean Square Error (RMSE), the coefficient of determination (R2 score), and the mean absolute error (MAE). These metrics are fundamental in quantifying the precision and explanatory power of DL models, providing a balanced view of their performance.  The RMSE measures the average magnitude of the model errors in the prediction of quantitative data. It provides a quantifiable measure of error magnitude that is directly relevant to the clinical application of ECG analysis, where accurate measurement is critical for patient diagnosis and treatment planning. A lower RMSE value indicates better performance, providing a clear measure of how the regressed values vary from the actual data points. In contrast, the R2 score assesses the proportion of variance in the dependent variable, which can be estimated from the independent variables. A higher R2 score suggests a better model explanatory power. R2 score complements the RMSE and helps to estimate the interpretable power of the models. The MAE is used to compare the designed models with state-of-the-art ECG parameter regression methods. Lower MAE represents better model performance. These metrics are useful for ECG analysis, as they assess the regression capabilities of the models used to explain the variations in the ECG parameters. In a medical context, these matrices are important because of the accurate regression of the ECG parameters that directly affect patients.

\begin{figure}[ht]
    \centering
    \includegraphics[width=3.3in,height=3.3in,clip,keepaspectratio]{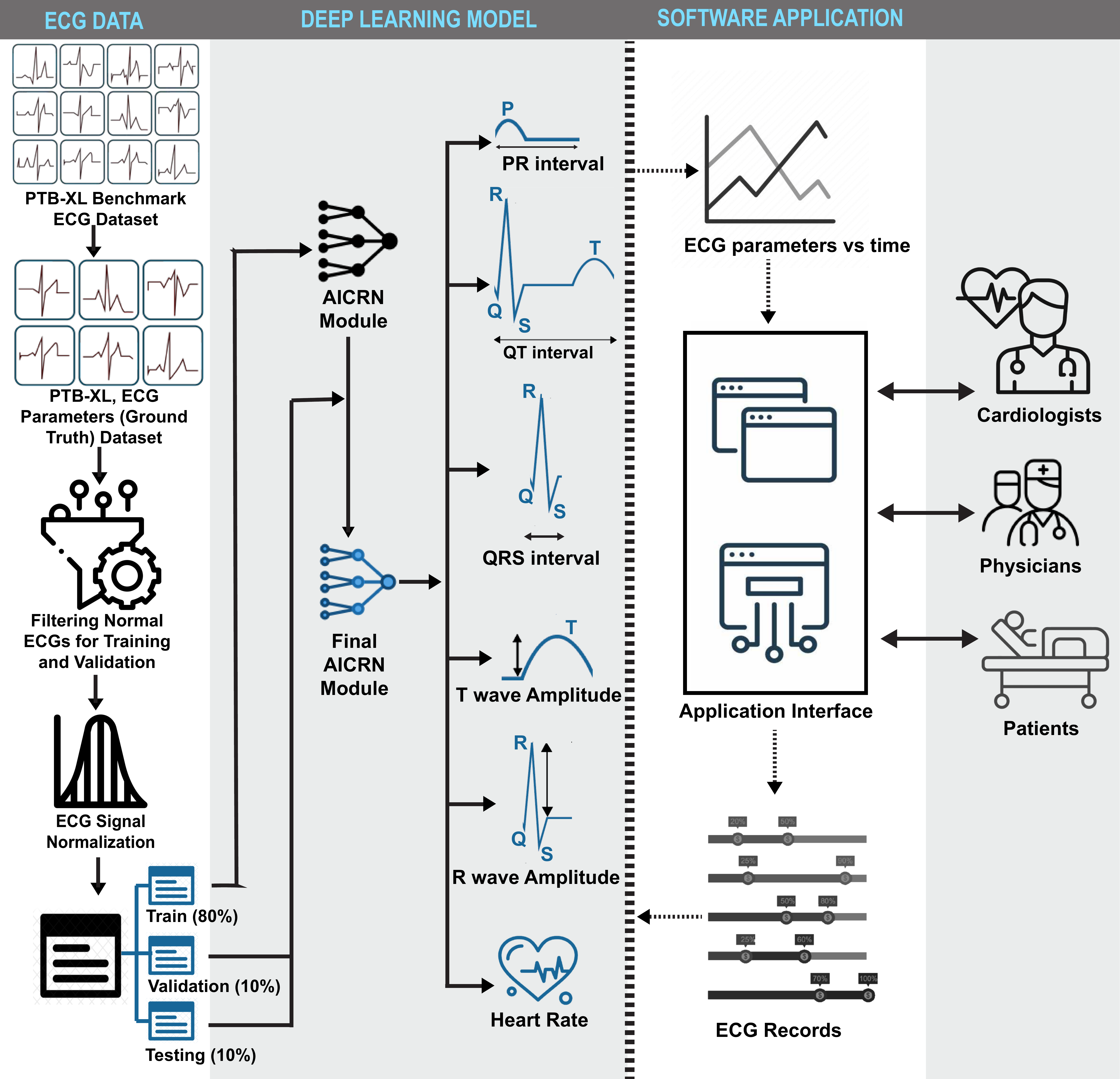}
    \caption{The system that uses six DL models for ECG analysis. It starts with ECG data sources, specifically mentioning the PTB-XL Benchmark ECG Dataset. Normal ECGs are filtered for training and validation. ECG signals are normalized. The models use a Convolutional Block Attention Resnet Module, and the data flows through multiple layers where various ECG parameters like PR interval, QT interval, QRS duration, heart rate, R-wave-peak amplitude, and T-wave amplitude are analyzed. The processed ECG records are input into the software application. The application plots ECG parameters over time. It is designed for three main user categories: physicians, cardiologists, and patients. The source code is available at \href{https://github.com/cepdnaclk/e17-4yp-Comprehensive-ECG-analysis-with-Deep-Learning-on-GPU-accelerators}{our GitHub repository}.
    }
    \label{fig:sa}
\end{figure}

\subsection{Application and Implications}
Manual Analysis of ECG Parameters and their interpretation requires significant expertise and takes considerable time. It has limitations in cardiac diagnosis for less experienced professionals. To overcome these challenges, this study developed an application using advanced DL models to automatically monitor and analyze patient ECG parameters, as shown in Figure~\ref{fig:sa}. It incorporates a long-term regression analysis of six ECG parameters (PR interval, QT interval, QRS complex, heart rate, T wave amplitude, and R wave peak amplitude) over consecutive months to make the proper decisions. The method can be used to quantitatively monitor the response to treatment in patients who take treatments according to the plans of medical professionals. The application offers automatic ECG analysis, which is possible even without specialist intervention. This technology optimizes the efficiency of specialists and increases their participation in cardiac care. This will enable medical professionals to make informed patient care decisions. The use of artificial intelligence in this context not only streamlines the diagnostic process but also enhances the accuracy of ECG interpretations.

\subsection{Tools and libraries}
Python served as the primary programming language, facilitating scripting and automation of data preprocessing tasks. Pandas is used for data manipulation and analysis in the data preprocessing stage, enabling efficient handling and transformation of large datasets. In addition, NumPy is used to perform complex numerical operations, which are essential in processing ECG data. Scripts import essential modules such as Pytorch for DL. These tools were chosen for their robustness and wide adoption, ensuring reproducibility and precise replication of the experimental setup. The models were trained on a selected device (NVIDIA Tesla C2050 / C2075), and the NAdam optimizer was initialized as an adaptive learning rate optimizer. It merged the root mean square propagation and Adam elements, and the mean square error loss function was defined for regression tasks.

\begin{table}[ht]
\centering
\caption{Results comparison for PR interval regression by the AICRN with other state-of-the-art architectures using mean absolute error. MEM \cite{b44} reconstructs the full ECG waveform from mobile SCG data and extracts key intervals, but this approach yields higher MAE than directly predicting intervals using the eight leads employed by AICRN.}
\label{table:deep_learning_ecg_performance_1}
\begin{tabular}{|c|ccc|c|}
\hline
\textbf{Output} & \multicolumn{3}{c|}{\textbf{Dataset}} & \textbf{PR (ms)} \\ \cline{2-4}
                & \textbf{Train} & \textbf{Validate} & \textbf{Test} & \\ \hline
AICRN           & PTB-XL            & PTB-XL              & -           & \textbf{4.620} \\ \hline
IKres \cite{b41}& MGH            & MGH              & PTB-XL           & 8.600  \\ \hline
MEM \cite{b44}  & Mobile data       & -        & Mobile data      & 12.020 \\ \hline
\end{tabular}
\end{table}

\begin{table}[ht]
\centering
\caption{Results comparison for QT interval regression by the AICRN with other state-of-the-art architectures using mean absolute error. Since the MGH dataset includes only 3 ECG leads information compared to the 8 relevant leads from PTB-XL used in the AICRN model, comparing their performance highlights the importance and benefit of leveraging comprehensive ECG lead information in deep learning-based ECG parameter regression. MEM \cite{b44} reconstructs the full ECG waveform from mobile SCG data and extracts key intervals, but this approach yields higher MAE than directly predicting intervals using the eight leads employed by AICRN.}
\label{table:deep_learning_ecg_performance_2}
\begin{tabular}{|c|c|c|c|c|}
\hline
\textbf{Output} & \multicolumn{3}{c|}{\textbf{Dataset}} & \textbf{QT (ms)} \\ \cline{2-4}
                & \textbf{Training} & \textbf{Validation} & \textbf{Testing} & \\ \hline
AICRN           & PTB-XL            & PTB-XL              & -           & \textbf{4.583} \\ \hline
IKres \cite{b41}& MGH            & MGH              & PTB-XL           & 10.700 \\ \hline
QTNet \cite{b42}& MGH               & MGH                 & MGH                & 12.630  \\ \hline
QTNet2 \cite{b43}& MGH              & BWH                 & MGH          & 12.000   \\ \hline
MEM \cite{b44}  & Mobile data       & -         & Mobile data      & 16.640 \\ \hline
\end{tabular}
\end{table}

\section{results comparison}
This study delves into the performance comparison between various neural network architectures, with the designed AICRN architecture in the regression of ECG parameters as shown in Table~\ref{table:deep_learning_ecg_performance_1}, Table~\ref{table:deep_learning_ecg_performance_2}, Table~\ref{table:deep_learning_ecg_performance_3}, Table~\ref{table:deep_learning_ecg_performance_4}, Table~\ref{table:deep_learning_ecg_performance_5} and Table~\ref{table:deep_learning_ecg_performance_6}. AICRN combines attention mechanisms with the CRN structure, enhancing the accuracy of ECG parameter estimations. These attention mechanisms allow models to prioritize the most relevant information from the ECG signals, which is critical for precision. In addition, the incorporation of residual connections addresses the problem of the vanishing gradient, enabling the effective training of deeper networks. This approach achieves superior performance.

Compared to AICRN, the IKres (2024) model \cite{b41}, based on a simpler ResNet architecture, lacks the specialized enhancements seen in AICRN, showing MAE differences of 3.980 ms in PR, 6.117 ms in QT, 4.392 ms in QRS, and 0.652 bpm in heart rate regression. In contrast, both QTNet (2023) \cite{b42} and QTNet2 (2024) \cite{b43} show differences of 8.047 ms and 7.417 ms in MAE. Despite their specialization in QT interval estimation, they do not compensate for the lack of broader capability enhancements, which AICRN’s attention-enhanced ResNet provides with 8 leads. These results imply that AICRN is configured with fewer layers of residual networks that do not optimize as effectively for the characteristics of the ECG signal. However, this shows the importance of increasing ResNet modules, as in the ResNet-18 architecture.

\begin{table}[ht]
\centering
\caption{Results comparison for QRS duration regression by the AICRN with other state-of-the-art architectures using mean absolute error. Since the MGH dataset includes only 3 ECG leads compared to the 8 relevant leads from PTB-XL used in the AICRN model, comparing their performance highlights the importance and benefit of leveraging comprehensive ECG lead information in deep learning-based ECG parameter regression. MEM \cite{b44} reconstructs the full ECG waveform from mobile SCG data and extracts key intervals, but this approach yields higher MAE than directly predicting intervals using the eight leads employed by AICRN.}
\label{table:deep_learning_ecg_performance_3}
\begin{tabular}{|c|c|c|c|c|}
\hline
\textbf{Output} & \multicolumn{3}{c|}{\textbf{Dataset}} & \textbf{QRS (ms)} \\ \cline{2-4}
                & \textbf{Training} & \textbf{Validation} & \textbf{Testing} & \\ \hline
AICRN           & PTB-XL            & PTB-XL              & -           & \textbf{2.008} \\ \hline
IKres \cite{b41}& MGH            & MGH              & PTB-XL           & 6.400 \\ \hline
MEM \cite{b44}  & Mobile data       & -         & Mobile data      & 16.690 \\ \hline
\end{tabular}
\end{table}

\begin{table}[ht]
\centering
\caption{Results comparison for heart rate regression by the AICRN with other state-of-the-art architectures using mean absolute error. Since the MGH dataset includes only 3 ECG leads compared to the 8 relevant leads from PTB-XL used in the AICRN model, comparing their performance highlights the importance and benefit of leveraging comprehensive ECG lead information in deep learning-based ECG parameter regression. MEM \cite{b44} reconstructs the full ECG waveform from mobile SCG data and extracts key intervals, but this approach yields higher MAE than directly predicting intervals using the eight leads employed by AICRN.}
\label{table:deep_learning_ecg_performance_4}
\begin{tabular}{|c|c|c|c|c|}
\hline
\textbf{Output} & \multicolumn{3}{c|}{\textbf{Dataset}} & \textbf{HR (bpm)} \\ \cline{2-4}
                & \textbf{Training} & \textbf{Validation} & \textbf{Testing} & \\ \hline
AICRN           & PTB-XL            & PTB-XL              & -           & \textbf{0.428} \\ \hline
IKres \cite{b41}& MGH            & MGH              & PTB-XL           & 1.080 \\ \hline
QTNet2 \cite{b43}& MGH              & BWH                 & MGH          & 1.200 \\ \hline
MEM \cite{b44}  & Mobile data       & -         & Mobile data      & 1.840 \\ \hline
\end{tabular}
\end{table}

\begin{table}[ht]
\centering
\caption{Results comparison for R wave peak amplitude regression by the AICRN with other state-of-the-art architectures using mean absolute error.}
\label{table:deep_learning_ecg_performance_5}
\begin{tabular}{|c|c|c|c|c|}
\hline
\textbf{Output} & \multicolumn{3}{c|}{\textbf{Dataset}} & \textbf{RPA (mV)} \\ \cline{2-4}
                & \textbf{Training} & \textbf{Validation} & \textbf{Testing} & \\ \hline
AICRN           & PTB-XL            & PTB-XL              & -           & \textbf{0.027}  \\ \hline
LeNet \cite{b36}& PTB-XL            & PTB-XL              & -           & 0.060 \\ \hline
XResNet \cite{b36}& PTB-XL          & PTB-XL              & -           & 0.080 \\ \hline
\end{tabular}
\end{table}

\begin{table}[ht]
\centering
\caption{Results comparison for T wave amplitude regression by the AICRN with other state-of-the-art architectures using mean absolute error.}
\label{table:deep_learning_ecg_performance_6}
\begin{tabular}{|c|c|c|c|c|}
\hline
\textbf{Output} & \multicolumn{3}{c|}{\textbf{Dataset}} & \textbf{TWA (mV)} \\ \cline{2-4}
                & \textbf{Training} & \textbf{Validation} & \textbf{Testing} & \\ \hline
AICRN           & PTB-XL            & PTB-XL              & -           & \textbf{0.028} \\ \hline
LeNet \cite{b36}& PTB-XL            & PTB-XL              & -           & 0.035 \\ \hline
XResNet \cite{b36}& PTB-XL          & PTB-XL              & -           & 0.038 \\ \hline
\end{tabular}
\end{table}

MEM (2023) \cite{b44}, designed for ECG analysis using mobile phone sensors, could introduce variability due to less controlled acquisition environments. Its encoder-decoder architecture, though sophisticated, struggles with the variability and lower quality of data from mobile sensors compared to the robust, attention-focused analysis, showing 7.400 ms in PR, 12.057 ms in QT, 14.682 ms in QRS, and 1.412 bpm in heart rate MAE differences in regression. Similarly, QTGAN (2024) \cite{b45} employs a Generative Adversarial Network approach to estimate QT intervals. The bias in the estimate of the QT interval was \SI{-7.1}{ms} with an upper limit of 50.170 ms and a lower limit of \SI{-64.380}{ms}. Furthermore, these results emphasize the importance of integrating generative adversarial networks into AICRN architecture to enhance the performance.

The comprehensive handling of multiple ECG characteristics through the attention-integrated approach allows consistently strong performance in the peak amplitude of the R wave and the peak amplitude of the T wave. This is confirmed by 0.033 mv in RPA and 0.007 mv in TWA MAE differences by LeNet (2024) \cite{b36} and 0.053 mv in RPA and 0.010 mv in TWA MAE differences by XResNet (2024) \cite{b36}. This architectural integration allows it to outperform other models. Attention mechanisms significantly reduce error by focusing on the most informative parts of the ECG signal, leading to accurate and reliable estimations of ECG parameters under various conditions.

\begin{table}[ht]
\centering
\caption{Comparison of performance metrics (RMSE and $R^2$ score) from 5 runs of attention-integrated and non-attention residual deep learning models for ECG parameter estimation using the PTB-XL dataset. Each entry presents the mean ± standard deviation.}
\label{table:ablation}
\begin{tabular}{|p{1.5cm}|c|c|c|c|}
\hline
\textbf{Parameter} & \textbf{Metric} & \textbf{Including Attention} & \textbf{Without Attention} \\ \hline
\multirow{2}{*}{PR (ms)} & RMSE & \textbf{5.047} ± 0.687 & 5.343 ± \textbf{0.286} \\
                         & $R^2$ & \textbf{0.964} ± 0.010 & 0.941 ± \textbf{0.008} \\ \hline
\multirow{2}{*}{QT (ms)} & RMSE & \textbf{4.614} ± \textbf{0.288} & 5.108 ± 0.378 \\
                         & $R^2$ & \textbf{0.976} ± \textbf{0.001} & 0.970 ± 0.004 \\ \hline
\multirow{2}{*}{QRS (ms)} & RMSE & \textbf{2.379} ± 0.267 & 2.846 ± \textbf{0.219} \\
                          & $R^2$ & \textbf{0.936} ± 0.011 & 0.900 ± \textbf{0.015} \\ \hline
\multirow{2}{*}{HR (BPM)} & RMSE & \textbf{0.473} ± \textbf{0.043} & 0.606 ± 0.122 \\
                          & $R^2$ & \textbf{0.998} ± \textbf{0.0001} & 0.997 ± 0.0009 \\ \hline
\multirow{2}{*}{RPA (mV)} & RMSE & \textbf{0.044} ± 0.004 & 0.053 ± \textbf{0.002} \\
                          & $R^2$ & \textbf{0.989} ± 0.003 & 0.985 ± \textbf{0.002} \\ \hline
\multirow{2}{*}{TWA (mV)} & RMSE & \textbf{0.031} ± 0.004 & 0.032 ± \textbf{0.001} \\
                          & $R^2$ & \textbf{0.961} ± 0.005 & 0.951 ± \textbf{0.002} \\ \hline
\end{tabular}
\end{table}

\section{Ablation study}
This study performed an ablation test comparing two versions of neural networks designed to analyze and estimate ECG parameters, as shown in Table~\ref{table:ablation}: one with attention mechanisms and the other without attention mechanisms. The version with attention mechanisms, called AICRN, consistently outperformed the basic convolutional residual network (CRN) in terms of mean. However, the residual network architecture without attention mechanisms can observe low variance in PR, QRS, RPA, and TWA. This highlights the importance of improving the AICRN's attention mechanisms according to the selected parameters. AICRN demonstrated low variance in QT and heart rate measurements, showing superior performance. In general, the attention mechanism sharpened the focus of the network on key ECG features, improving performance and consistency.

\section{Discussion}
The major discovery of the research is the development of an AICRN architecture that dramatically outperforms existing methods, parallel with the accuracy of ECG parameters important in regression. The explanatory power improved further, as attention mechanisms were introduced into a residual network. In the PTB-XL dataset, AICRN achieved low error rates and high explanatory power, clearly showing strong clinical reliability. In addition, the trained models have the potential to be integrated into real-time applications for automated ECG monitoring. The software application designed using the models provides an automated ECG regression and a reduced time duration for professional ECG analysis.

Despite the excellent performance of AICRN employed in parameter regression, there were some discrepancies and unexpected results. In the ablation study, the performance variances between AICRN (with attention) and CRN (without attention) were smaller for the PR interval, QRS duration, R-wave peak amplitude, and T-wave amplitude, suggesting a limitation of the attention mechanisms. Additionally, under some parameter criteria (such as heart rate), simple baseline model architectures, such as regular residual networks without attention mechanisms, performed well. It was found that AICRN often outperformed in regression variance. This comes at the cost of questions in terms of complex attention mechanisms that may come into play. In addition, previous studies demonstrated that transformers were preferable in ECG analysis due to their ability to detect long-range dependencies. However, the study’s well-optimized AICRN excelled in ECG parameter regression, showing the value of simple, efficient architectures.

Although our AICRN significantly improves the accuracy of ECG analysis, its performance depends on the diversity and quality of the training data. This highlights the importance of training on varied datasets to ensure robust performance in different clinical settings. DL models often require large datasets to learn effectively. However, rare cardiac events, which are critically important for diagnosis, are not adequately represented in available datasets. This can limit the models' ability to detect and diagnose rare but severe conditions accurately. Although the models achieve high accuracy, interpretability remains a challenge. Models must be transparent to gain the trust of medical practitioners and must be integrated effectively into clinical practice. The deployment of AI models in healthcare settings is subject to stringent regulatory requirements to ensure safety and efficacy. Obtaining compliance can be challenging and time-consuming, potentially delaying the implementation of innovative AI solutions in clinical practice.

The advancement of AICRNs in ECG analysis is vital for identifying various cardiac conditions, including the increased risks of arrhythmias and conduction abnormalities. Our models leverage advanced neural network architectures to enhance the precision of the regression of ECG parameters. By continuously analyzing patient ECG data, this system can regress minor changes that could elude human experts, providing a comprehensive assessment of cardiac health. This level of detail ensures that even subtle anomalies are caught early, facilitating interventions that can prevent serious conditions from developing. This application is crucial for monitoring cardiac functions, enabling accurate predictions of various ECG parameters. These model capabilities support intraoperative monitoring, preoperative heart evaluation, and assessment of blood electrolyte levels.

The ability of the application to make timely and accurate diagnoses is crucial for initiating early treatments and monitoring responses to medications, thus reducing adverse effects and costs. This proactive approach to healthcare delivery not only improves patient outcomes but also optimizes resource allocation within medical facilities. Furthermore, the application's predictive capabilities, driven by its advanced neural network architecture, enable the forecasting of potential cardiac events. It also enables medical professionals to efficiently monitor a larger number of patients with less specialized expertise, improving healthcare delivery and accessibility.

By integrating similar applications, healthcare providers can identify and manage emerging issues before they become critical with less expertise and knowledge. Integrating such applications into healthcare systems reduces the workload on healthcare professionals by automating routine monitoring and analysis tasks. This not only improves healthcare delivery efficiency but also enables cardiologists to focus on complex cases and patient interactions. This application is particularly beneficial for patients with chronic heart disease and for those in remote areas with limited access to cardiologists and advanced diagnostic tools. Decentralizing the availability of high-quality cardiac care significantly improves the accessibility and equity of healthcare services.

AICRN is a novel direct regression framework for ECG parameters. Although most previous works have used only classification or segment-level detection strategies for ECG analysis, this study presents a new end-to-end regression method for the estimation of clinically relevant intervals and amplitudes using the AICRN architecture. Examining the large PTB-XL dataset and applying the models in applications, this work demonstrates the ability to interpret and automate ECG analysis and overcome the shortcomings in model precision and clinical applications observed in previous work.

To extend the horizon of use and further optimize the system, there are several approaches to advanced DL-based ECG analysis, such as adding convolutional residual network layers and applying powerful data augmentation methods such as GANs. Such networks create diverse synthetic ECG signals that train DL to rapidly recognize rare cardiac conditions. Noise reduction algorithms are needed in ECG signals. Research through wide-ranging data sets that span human populations would not only help to ensure that the models were effective but also unbiased in terms of their population proportion. Extending this system for explainability promotes transparency, increasing clinician trust and understanding.

The application has the potential to improve patient participation and promote self-management. The designed system can be improved to facilitate immediate medical-level feedback on cardiac health, motivating them to participate in treatment plans and lifestyle changes. Furthermore, the development of multimodal DL approaches for ECG analysis broadened the scope from traditional cardiac diagnosis to systems capable of identifying emotional states or biometric identification. They also enable the storage, retrieval, and sharing of ECG data, facilitating remote consultations and long-term monitoring. More research is needed on low-power and efficient deep learning models for wearable ECG devices to enable real-time cardiovascular monitoring in outpatient and remote settings.

\section{Conclusions}
The main result of this research is the development of an AICRN architecture that dramatically outperforms existing methods with respect to the MAE of regressing important ECG parameters. As attention mechanisms are introduced into a residual network, the interpretable power is further improved by regression precision. In the PTB-XL dataset, AICRN achieved low error rates, implying high interpretability power, clearly showing strong clinical reliability. The designed neural network architecture can be integrated into real-time and explainable cardiac monitoring systems. In addition, we developed an application for automated ECG monitoring that provides an excellent rate of parameter regression and cardiac care for patients, which is important in limited specialist settings. The stability of our models is well evidenced by extensive experiments with the PTB-XL dataset, with lower MAEs for all ECG parameters compared to ResNet, QTNet, and QTGAN. Our AICRN channel and spatial attention architecture outperformed the baseline architecture in mean and variance of the regression results by prioritizing the ECG characteristics. The designed software application opens the doors to a rich and user-friendly set of automated cardiac monitoring applications based on our models in high patient load and resource-constrained conditions, opening up new avenues for accurate, instant cardiac care and reduced variability in human diagnosis. Generalizing the findings requires further validation across diverse datasets to ensure model effectiveness across populations and clinical settings. Furthermore, while our models achieve high accuracy, we also notice ethical considerations related to their application in clinical practice (patient privacy, interpretability of machine decisions, etc.). Clinical studies are crucial in evaluating the real-world effectiveness of the designed architecture. The interpretability of models is an important prerequisite for the future development of models to ensure transparency and reliability in clinical decisions. The techniques for practitioners to understand and validate algorithms are crucial for the trustworthiness and successful adoption of AI tools in medical practice. The proposed architecture will help design explainable modules that can help extend and improve interpretation and transparency in cardiac diagnosis using artificial intelligence techniques. In this regard, our study provides a platform for future research and development that will lead to significant improvements in the prognosis of diseases and the response to cardiac disease treatment, and therefore patient outcomes through better decision-making.

\begin{IEEEbiography}[{\includegraphics[width=1in,height=1.25in,clip,keepaspectratio]{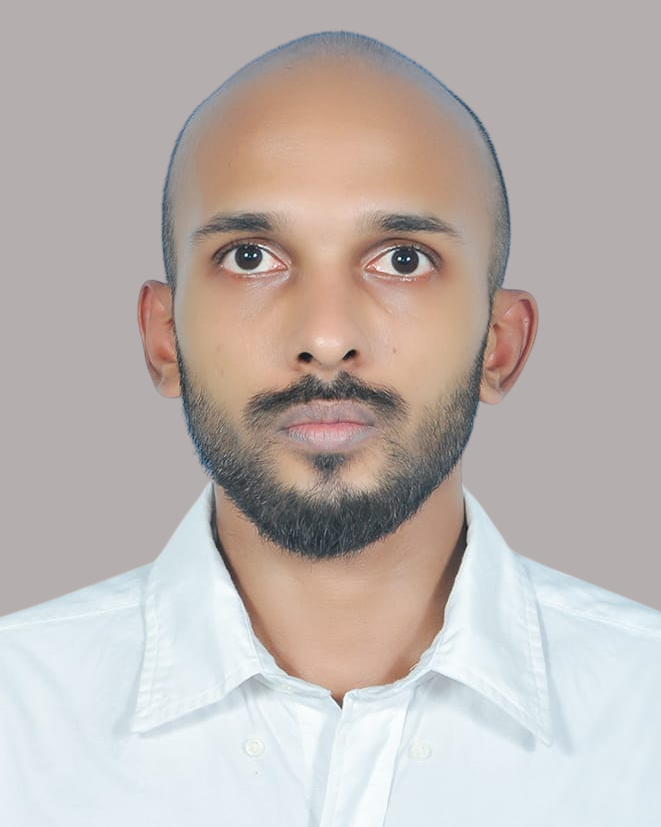}}]{J. M. I. H. Jayakody} received his B.Sc. (Hons.) in Computer Engineering from the University of Peradeniya, Sri Lanka, in 2025. He currently serves as a Teaching Assistant in the Department of Computer Engineering, actively involved in academic and research work.

His research focuses on randomness, deep learning, applied mathematics, and algorithms to develop efficient and scalable solutions that explore the intersection of traditional and modern computational techniques to address complex engineering challenges.
\end{IEEEbiography}

\begin{IEEEbiography}[{\includegraphics[width=1in,height=1.25in,clip,keepaspectratio]{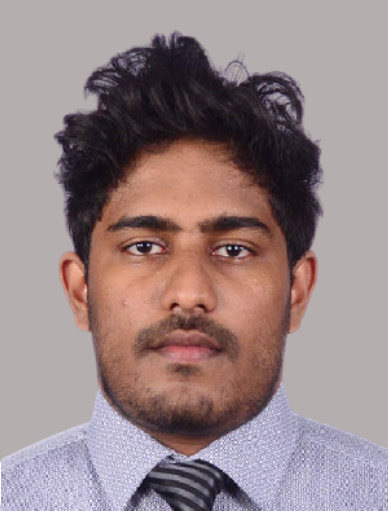}}]{A. M. H. H. ALAHAKOON} received the B.Sc. (Eng.) degree (Hons.) in computer engineering from the University of Peradeniya, Sri Lanka, in 2023. He is involved in research related to deep learning, image processing, and computer architecture. Currently a PhD candidate at the University of Sydney School of Electrical and Computer Engineering, focusing on research in making hardware-efficient AI.
\end{IEEEbiography}

\begin{IEEEbiography}[{\includegraphics[width=1in,height=1.25in,clip,keepaspectratio]{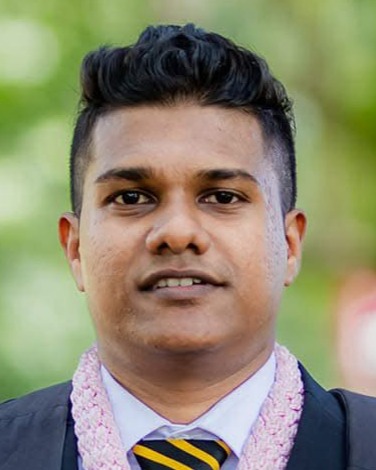}}]{C.R.M. Perera} received his B.Sc. Engineering (Hons) degree in Computer Engineering from the University of Peradeniya, Sri Lanka.

His research interests include machine learning, deep learning, artificial intelligence, computer hardware, and cloud systems. He is particularly interested in the integration of intelligent algorithms with hardware-aware systems and scalable cloud-based architectures. 
\end{IEEEbiography}

\begin{IEEEbiography}[{\includegraphics[width=1in,height=1.25in,clip,keepaspectratio]{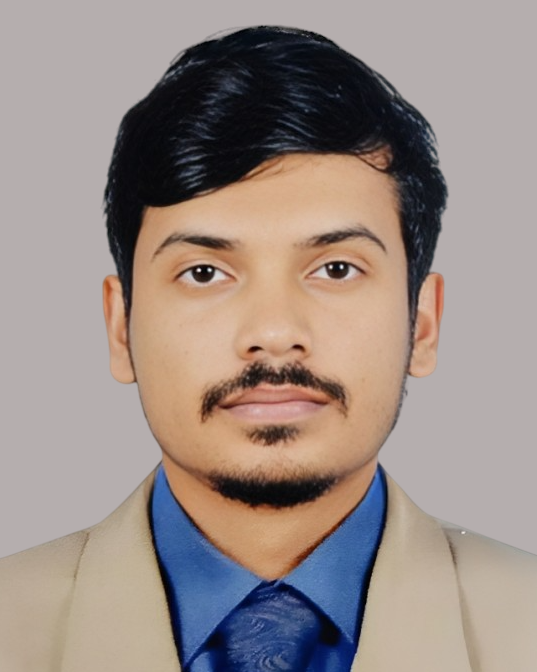}}]{R.M.L.C. Srimal} recieved his B.Sc. Eng (Hons) in Computer Engineering from the University of Peradeniya in 2023.

 His research examines the role AI can play in solving complex problems which include but are not limited to automation and data-centric decision-making. With cutting-edge technological research, he intends to design intelligent systems that enhance performance and practical success in critical areas.

\end{IEEEbiography}

\begin{IEEEbiography}[{\includegraphics[width=1in,height=1.25in,clip,keepaspectratio]{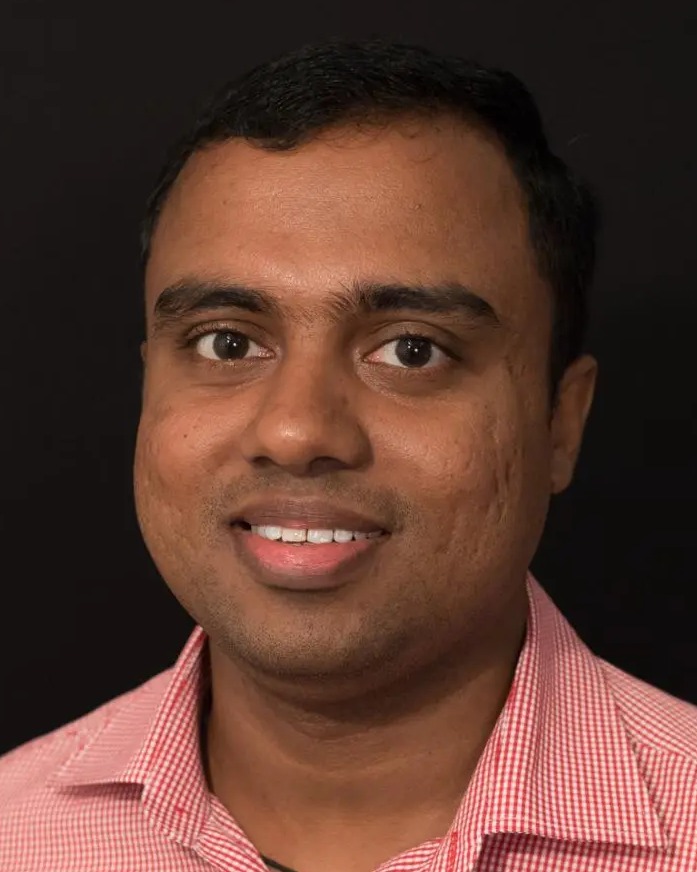}}]{Vajira Thambawita} (Member, IEEE) received his Ph.D. from Oslo Metropolitan University, Norway, in 2021. He is currently a Senior Research Scientist at the Simula Metropolitan Center for Digital Engineering (SimulaMet) and an Adjunct Associate Professor at Oslo Metropolitan University.

His research focuses on machine learning and deep learning, particularly in medical imaging and multimedia data analysis. He has made significant contributions to medical diagnostics, including polyp detection, segmentation, and the generation of synthetic datasets using GANs and diffusion models. Vajira is also recognized for his work on improving the generalizability of AI models. His interdisciplinary approach aims to develop AI-driven solutions that improve healthcare outcomes and address real-world challenges.
\end{IEEEbiography}

\begin{IEEEbiography}[{\includegraphics[width=1in,height=1.25in,clip,keepaspectratio]{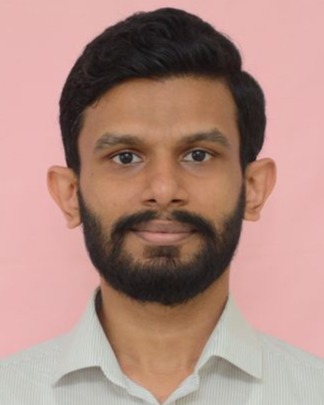}}]{Isuru Nawinne} received the B.Sc. degree in engineering from the University of Peradeniya,
Sri Lanka, in 2011, and the Ph.D. degree in computer science and engineering from The University of New South Wales, Sydney, Australia, in 2016. He is currently a Senior Lecturer in Computer Engineering with the University of Peradeniya. He has a keen interest in improving the quality of engineering education and is involved in curriculum development and the design of learner-centered delivery methods. His research interests include computer architecture, embedded systems,
biomedical engineering, and automation.

\end{IEEEbiography}

\begin{IEEEbiography}[{\includegraphics[width=1in,height=1.25in,clip,keepaspectratio]{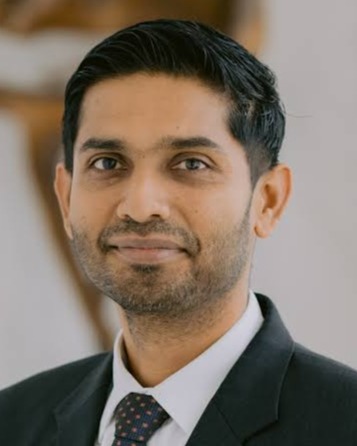}}]{Roshan Ragel} received his Ph. D. degree in Computer Science and Engineering from UNSW Sydney. He is a Professor at the Department of Computer Engineering of the University of Peradeniya, Sri Lanka. His research interest lies in systems-on-chip, IoT, high-performance computing, computational biology, artificial intelligence and wearable computing. He is the co-author or co-editor of over 200 peer-reviewed papers and severally honored: Elsevier Most Prolific Sri Lankan Author Award in 2017; Presidential Award for Scientific Publications (2018) and 2023; Sri Lanka's number 1 computer scientist by the AD Scientific Index (2021–2025).

In 2025 he was the recipient of the IEEE Computer Society's Mary Kenneth Keller Undergraduate Teaching Award (the first from the Global South) and of the 2024 National Educator Gold Award in Computing.

consulting CEO of LEARN (Sri Lanka's NREN) since 2017 a pivotal role in launching the Asi@Connect Project and also representing Sri Lanka in the governance of APAN and Asi@Connect.

He has been associated nationally in higher education policy, AI strategy, and curriculum formulation through the Presidential Task Force for Education, UGC, ICTA, and the founding member of AI Forum for Academics (AIFA).

\end{IEEEbiography}

\EOD

\end{document}